# Exploring the Utilities of the Rationales from Large Language Models to Enhance Automated Essay Scoring

Hong Jiao
University of Maryland, College Park
Hanna Choi
University of Maryland, College Park
Haowei Hua
Princeton University

## Abstract

This study explored the utilities of rationales generated by GPT-4.1 and GPT-5 in automated scoring using Prompt 6 essays from the 2012 Kaggle ASAP data. Essay-based scoring was compared with rationale-based scoring. The study found in general essay-based scoring performed better than rationale-based scoring with higher Quadratic Weighted Kappa (QWK). However, rationale-based scoring led to higher scoring accuracy in terms of F1 scores for score 0 which had less representation due to class imbalance issues. The ensemble modeling of essay-based scoring models increased the scoring accuracy at both specific score levels and across all score levels. The ensemble modeling of essay-based scoring and each of the rationale-based scoring performed about the same. Further ensemble of essay-based scoring and both rationale-based scoring yielded the best scoring accuracy with QWK of 0.870 compared with 0.848 reported in literature.

## Introduction

Automated essay scoring methodology develops along with the advances in AI technology. Starting from the early supervised machine learning models based on engineered features (e.g., Mahana et al., 2012) to recent use of large language models (LLMs), the methods for automated essay scoring as demonstrated in Appendix A evolved with the advances in machine learning, deep learning, language models, and LLMs. Using automated scoring of Prompt 6 in the Automated Student Assessment Prize (ASAP) dataset from Kaggle, this study intends to explore the utility of rationales generated by LLMs in enhancing automated essay scoring.

For the ASAP Prompt 6, automated scoring models have been developed since 2012 after the Kaggle competition. The methods used in the competition demonstrated the state-of-the-art in 2012 which achieved the highest quadratic weighted kappa (QWK) of 0.814 in the private leaderboard and 0.801 in the public leaderboard. Since then, different researchers have developed different models to improve scoring accuracy. For example, Taghipouer and Ng (2016) explored the Long-Short Term Memory (LSTM) model and achieved a QWK of 0.813 while Liang et al (2017) used bidirectional LSTM and word embeddings from GloVe to achieve a QWK of 0.820. Dong et al (2017) developed attention-based recurrent convolutional neural network yielding a QWK value of 0.811. In the same year, Zhao et al (2017) developed a memory-augmented neural model and increased QWK to 0.830.

Since the emergence of the BERT model, multiple studies have explored the BERT model and its variants when they became available. Rodriguez et al (2019) trained the BERT base model and achieved a QWK of 0.805 for Prompt 6 while BERT ensembled with CCN yielded a higher QWK value of 0.819. Yang et al (2020) achieved a record high QWK of 0.847 using $R^2$ BERT model. Xue et al (2021) developed a hierarchical BERT-based transfer learning approach for multi-dimensional essay scoring and achieved QWK of 0.832 for fine-tuned BERT, surpassing attention-based recurrent convolutional neural network (0.811; Dong, et al., 2017) and memory-augmented neural model (0.830; Zhao et al., 2017). Ormerod et al (2021) reported the efficient ensemble model integrating multiple language models including BERT, Albert base and large, Electra small, Mobile BERT, Reformer, and yielded the highest QWK of 0.822 for Prompt 6. Ormerod (2022) explored the DeBERTa large model and was able to increase QWK to 0.834 for Prompt 6. Cho et al (2023) explored data augmentation and was able to boost QWK for Prompt 6 to 0.839, next to the best QWK of 0.847 reported in Yang et al (2020) up to that time point.

LLMs such as ChatGPT, Gemini and Claude have been explored for automated scoring (e.g., Bui & Barrot, 2024; Mizumoto & Eguchi, 2023; Parker et al., 2023; Yancey et al., 2023; Wijekumar et al., 2024). Lee et al (2024) prompted LLMs for zero-shot essay scoring via multi-trait specialization with ChatGPT yielding the highest QWK of 0.668 for Prompt 6, much lower than fine-tuned language model. Shermi (2024) explored GPT-4o and obtained a QWK of 0.78 for Prompt 6. Jang et al (2025) explored ChatGPT and Claude for ASAP essay scoring and yielded QWK of 0.606 and 0.630 for Prompt 6 respectively. Shibata and Miyamura (2025) explored pairwise comparison using multiple LLMs for essay scoring. Their proposed approach increased QWKs by 0.2 to 0.4 for Mistral-7B, Llama 3.2-3B, Llama 3.1-8B, GPT-4o, and GPT-4o-mini, with Mistral-7B yielding the highest QWK of 0.792. Their further exploration of transductive and inductive settings in pairwise comparison in GPT-4o-mini for essay scoring increased QWK from 0.737 to 0.783 based on inductive comparison.

Thanks to their strong reasoning capabilities, LLMs can generate rationales to support the scores they assign. Yancey et al. (2023) demonstrated that requiring rationale generation has improved the LLM's zero-shot scoring performance. Lee et al. (2024) explored GPT-3.5, Llama2, and Mistral-7B for rationale generation with zero-shot prompting for multi-trait analytic scoring. Though the QWKs achieved by these LLMs for Prompt 6 were in the range of 0.5 to 0.67, their study demonstrated that generating rationales enhanced the transparency of automated essay scoring. It is expected that the information captured by rationales should explicitly reflect the direct evaluation of the key elements in the scoring rubrics. Xiao et al (2024) demonstrated that LLM feedback helps to improve GPT-4, Novice, and Expert scoring accuracy. They fine-tuned GPT-3.5 and achieved a QWK of 0.848, almost as high as the QWK achieved by Yang et al (2020) in training BERT for Prompt 6.

A few recent studies (Chu et al., 2024; Li et al., 2023) have explored the use of rationales in automated essay scoring. Li et al. (2023) instructed ChatGPT for rationale generation and automated holistic scoring for Science and Biology items. Their study used the rationales generated by ChatGPT to predict student short-answer scores and reported QWK ranging from .90 to 0.99 across all four items. They further compared score prediction with and without rationales in addition to Question, Key Elements, Rubric, and student answer as input and found improvement in QWK ranging from the lowest of 0.03 (0.82 -> 0.86) to the highest of 0.30 (0.58 -> 0.88). Chu et al. (2024) explored GTP-3.5 and Llama-3.1 for rationale generation based on prompt engineering. Their comparison among score prediction using essays, or rationales from either GPT or Llama demonstrated that QWK was the highest for essay-based scoring while that using the rationale from GPT followed. They fine-tuned several encode-decoder language models including T5, Flan-T5, BART, Pegasus, and Longformer Encoder-Decoder (LED) including rationales for automated analytic scoring.

In general, T5 performed the best but only yielded QWK of 0.773. Though the findings about the added values of rationales in automated scoring differed in these studies, LLM generated rationales is an additional source of information as data augmentation to enhance essay scoring accuracy.

## Purposes of the Study

Given only one study (Chu et al., 2024) explicitly evaluated the utilities of rationales generated by LLMs (GPT-3.5 and Llama-3.1) for automated scoring of analytic scores for multiple traits in essay writing, this study explores the utilities of rationales generated by LLMs in enhancing accuracy of automated holistic scoring of essays. More specifically, the following two research questions are explored.
1. Compared with essay-based scoring, how does rationale-based scoring perform?
2. Whether ensemble of language models based on essays and rationales will improve essay scoring accuracy?

## Methods

### Data

This study primarily used essay data from the Automated Student Assessment Prize (ASAP; https://www.kaggle.com/competitions/asap-aes/data). The ASAP data features essays for 8 prompts, written by US students in Grades 7 through 10, hand scored by 2 human raters. This study used essays for Prompt 6 with holistic scores. Two raters scored each essay. A resolution score is provided by keeping the higher score in the two ratings. The number of essays for Prompt 6 is 1,800. Essay length ranges from 3 to 454 with an average essay length of 150. The essay length distribution is presented in Figure 1. The score ranges from 0 to 4. The score distribution is presented in Table 1. Few students got scores of 0 or 1, especially, for score 0, a relatively small sample size of 44 out of 1,800 essays may increase the challenges in scoring accuracy for this score category.

Figure 1. Essay Length Distribution

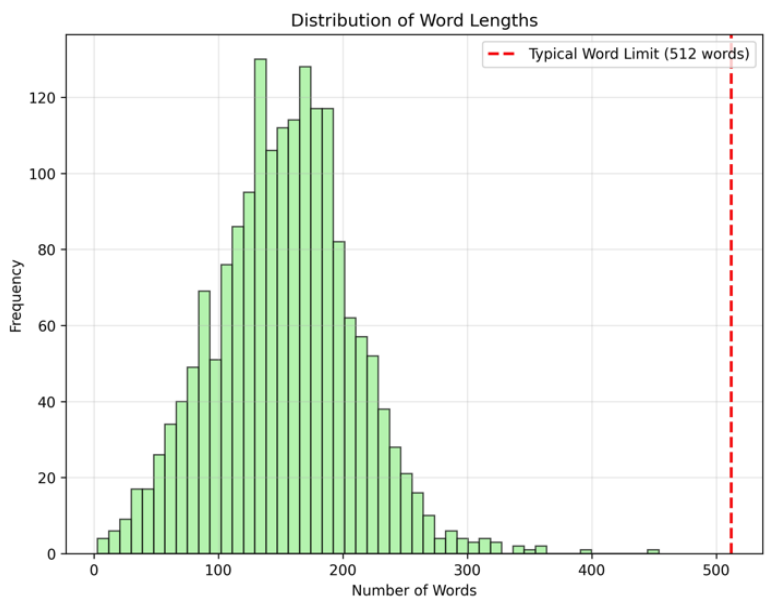

Table 1. Score Distribution for Prompt 6.

| Scores | Frequency | Percent |
|---|---|---|
| 0 | 44 | 2.4 |
| 1 | 167 | 9.3 |
| 2 | 405 | 22.5 |
| 3 | 817 | 45.4 |
| 4 | 367 | 20.4 |

**Study Design**
*Rationale Generation*

Two LLMs including GPT-4.1 and GPT-5 were explored for scoring and rationale generation. GPT 4.1 was chosen over GPT-4 and GPT-4o due to its reliability in complex, multi-step reasoning and relative cost efficiency. GPT-4.1 produces highest reasoning quality and dependable structured outputs. GPT-5 was chosen for its latest version of the GPT models.

The prompting template is zero-shot prompting with scoring rubric guidelines used to train human raters. As Prompt 6 is a passage dependent essay writing, the passage was included in prompting. The prompt for writing was included as well. A sample prompting is provided in Appendix B. LLM prompting, consisting of task, a passage, prompt for writing, scoring rubrics, additional notes to highlight the potential correct answers as provided in the documentation for Prompt 6, essays, and directions for outputs, essentially asked GPT to score each essay and provide rationale to justify the score assigned to each essay.

Both GPT-4.1and GPT-5 were used for scoring and rationale generation. API was used in batch run for scoring and rationale generation. The temperature was set at 0.2 for GPT-4.1 to reduce randomness in rationale generation while that for GPT-5 was set at 1 as the use of the API key for GPT-5 does not allow the change of temperature to a lower value than 1. GPT-4.1 used the prompting template described above. However, it was found for GPT-4.1 runs, some rationales were longer than 512 token length. Thus, when GPT-5 was used for scoring and rationale generation, additional prompting was added to highlight the use of succinct language to generate the rationales as the use of API keys did not allow to set the maximum token length for GPT-5. For scoring and rationale generation, GPT-4.1 used for about 206 minutes. However, GPT-5 used about 590 minutes to complete the tasks. GPT -4.1 cost about $24 while GPT-5 cost about $50 for 1800 essays.

*Scoring Model Development*

For model development, two sets of input data were explored for automated holistic scoring; one was based on essays while the other based on rationales from either GPT. As it is not clear whether the information extracted from essays and rationales are complementary to each other, this study first developed models using essays and rationales separately. As the essay lengths were within the reasonable range of token lengths for encoder-based language models (LMs), this study explored several LMs including BERT (Delvin et al., 2018), deBERTa (He et al., 2020), distilBERT (Sanh et al., 2019), ELECTRA (Clark et al., 2020), and RoBERTa (Liu et al., 2019). Only base models were included for BERT and distilBERT. For deBERTa and RoBERTa, both base and large models were explored. Only ELECTRA large model was trained. All these seven models were trained using essays or rationales as input data and model performance was evaluated for models based on essays and rationales from GPT-4.1 and GPT-5.

These common encoder-only Transformer models were chosen for their state-of-the-art (SOTA) for automated scoring. BERT as Encoder-only Transformer with learned absolute positional embeddings sums token, position, and token-type (segment) embeddings. It is pretrained with masked language modeling (MLM) and next sentence prediction (NSP) on BookCorpus + Wikipedia. It uses static masking. RoBERTa, the same basic encoder with learned absolute positions.as BERT but removes NSP, uses dynamic masking and much more data and training (CommonCrawl-like corpora). DeBERTa-v3 uses disentangled attention which separates content and position representations and relative position biases instead of adding absolute positions to inputs with improved decoding. DeBERTa v3 also uses replaced token detection (ELECTRA-style) in addition to MLM for efficiency and accuracy. It was trained on larger corpora. RoBERTa (base/large), DeBERTa base and DeBERTa-v3 large are strong SOTA for classification, ELECTRA is sample-efficient pretraining while DistilBERT is smaller and faster. For classification, it is expected that DeBERTa-v3 and RoBERTa generally outperform vanilla BERT.

In addition to the base encoder LMs, we also trained multiple ensemble models to improve scoring accuracy using the predicted scores based on essays and rationales. This is different from the study by Chu et al. (2024), where they included rationales as the input data for training models for long context including T5, Flan-T5, BART, Pegasus, and LED. We explored seven ensemble models. The first one is QWK Optimized Ensemble (Caruana et al., 2004; Perrone, & Cooper, 1992) which use the exponential weighting based on QWK scores using the equation: weight = exp(5 * (qwk - 0.8)). This weighting favors high-performing models more aggressively compared to simple linear weighting. This method is simple and interpretable while directly optimizing for QWK when it automatically downweights poor performers. However, it may overweight a few good models too aggressively and doesn't consider model correlations or interactions. The second is Elite Ensemble (Opitz & Maclin, 1999; Zhou, Wu, & Tang, 2002), which only uses top-performing models such as QWK > 0.8. It assigns fixed weights based on performance ranking such as Electra-large (essay) with 35% weight, Deberta-v3-large (essay) with 25% weight and other strong performers get smaller weights. It eliminates noise from weaker models and focuses computational resources on best predictors. By removing contradictory signals from poor performers and leveraging domain knowledge about which models are reliable, this method may work very well. Another method is Weighted Median Ensemble (Kittler et al., 1998). Instead of weighted average, this method uses weighted median which could be more robust to outliers than mean-based methods. This method is less sensitive to poorly calibrated models and maintains ordinal nature of scoring. The fourth is Confidence Weighted Ensemble (Niculescu-Mizil & Caruana, 2005; Platt, 1999). This method measures prediction confidence as in confidence = 1 - |prediction - rounded_prediction| when regressor head was used. The confidence is higher when predictions are closer to integer scores. This method only includes predictions above a confidence threshold (0.6) and squares confidence to favor high-confidence predictions. This approach is adaptive using models only when they're "confident". It reduces impact of uncertain predictions. However, confidence measures might not reflect true model uncertainty and may discard useful information from moderately confident predictions. The fifth ensemble is Tiered Ensemble (Fan et al., 2005) which makes score-range specific adjustments. It uses elite ensemble as base, applies special rules for extreme scores as follows. For scores < 1.0, it pulls predictions down slightly while for scores > 3.0, it pulls predictions up slightly. In this way, it makes conservative adjustments to extreme scores. This method addresses known scoring biases at extremes and incorporates domain knowledge about scoring tendencies. This method can deal with essay scoring biases at score boundaries and make conservative adjustments that prevent overconfident extreme predictions. The sixth method is Stacking Ensemble (Wolpert, 1992) which uses meta-learning based on predictions as features. It trains a Ridge regression model to combine predictions optimally, uses cross-validation to find best regularization parameter (alpha), and learns optimal linear combination weights. This method is data-driven, learning optimal weights from validation performance. The last method is Correlation Optimized Ensemble (Kuncheva & Whitaker, 2003) that combines QWK scores with correlation to actual scores with Weight = (QWK³) * (1 + correlation). It favors models that are both accurate and well-calibrated while considering both ranking (QWK) and calibration (correlation). It penalizes models with good QWK but poor correlation, leading to more comprehensive performance assessment. This method still uses fixed formula rather than learned weights and correlation might be noisy with limited data. It is expected that Stacking works best because it learns optimal combinations rather than using fixed rules. Tiered approach is excellent for incorporating domain knowledge about scoring patterns. Elite selection is a good simplification strategy that maintains performance while Robust methods (median, confidence) provide insurance against poor models but may limit peak performance.

*Model Training and Testing*

All models were trained and tested using the same dataset split of 80/20 between train/validation and test to ensure fair comparisons. The sample sizes for train, validation, and test datasets were 1260, 180, and 360 essays respectively. For GPT-based approaches, the same test dataset was used for model performance evaluation. Hyperparameters including learning rate (1e-5, 2e-5), epoch (10, 15), batch sizes (4, 8, 16) were tuned for better model performance.

*Model Performance Evaluation*

To determine model performance, QWK was computed using equation 1 to evaluate AI and human score agreements accounting for the severity of disagreement. Spearman rank order correlation was also computed. As the score distribution is not balanced, F1 for each score category was computed to examine the scoring accuracy within each score category. To compute F1 score, recall and precision were computed first. Recall (sensitivity, power, or true positive rate) is the metric that evaluates a model's ability to predict true positives out of the total actual positives, computed as in equation 2. Precision computes the proportion of actual positives out of the *predicted* positives. It is computed using equation 3. In order to seek the balance of both Recall and Precision, F1-scores (Forman & Scholz, 2010), the harmonic mean of recall and precision is computed for imbalanced class. According to Sasaki (2007), the F1 score is computed as in equation 4.

$$QWK = \frac{2 \times (TP \times TN - FN \times FP)}{(TP+FP) \times (FP+TN) + (TP+FN) \times (FN+TN)}, \quad (1)$$

$$Recall = \frac{TP}{TP+FN}, \quad (2)$$

$$Precision = \frac{TP}{TP+FP}, \quad (3)$$

$$F1 = \frac{2 * Precision * Recall}{Precision + Recall} = \frac{2 \times TP}{2 \times TP + FP + FN}, \quad (4)$$

where TP=true positive, FP=false positive, TN=true negative, and FN=false negative.

# Results

The word length distribution for rationales generated by GPT-4.1 and GPT 5 is summarized in Figures 2 and 3 respectively. The GPT-4.1 rationale length ranged from 103 to 504 with an average essay length of 310.3 while the GPT-5's rationale length ranged from 60 to 197 with an average rationale length of 129.5. The token sizes for the rationales generated by GPT-5 did not exceed the limit. However, the longer rationales generated by GPT-4.1 led to the number of tokens exceeding the limit of 512 for different LMs with an overage ranging from 113 to 189 rationales. Essay length only led to several essays with token sizes over the limit of 512. For essays or rationales with over 512 token sizes, truncation was applied automatically in model training.

Figure 2. Rationale Length Distribution for GPT-4.1

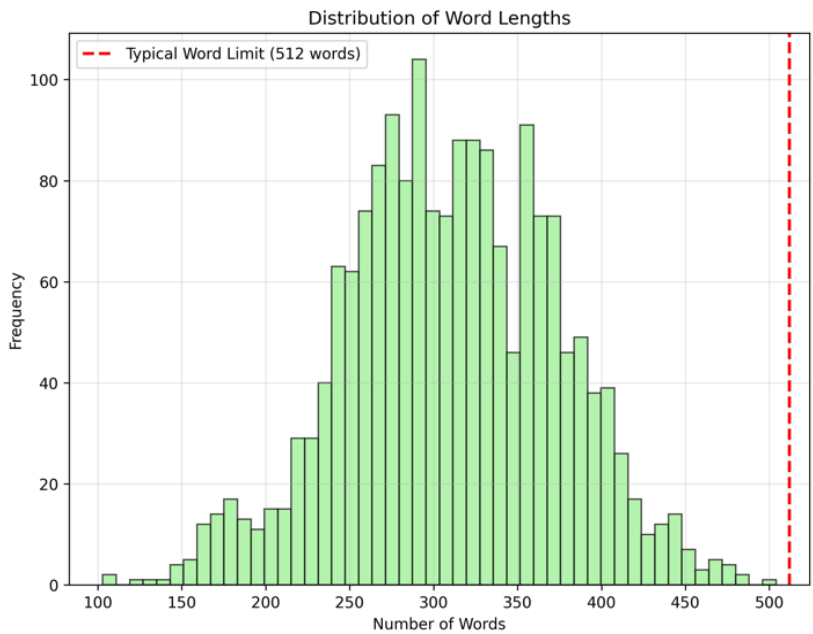

Figure 3. Rationale Length Distribution for GPT-5

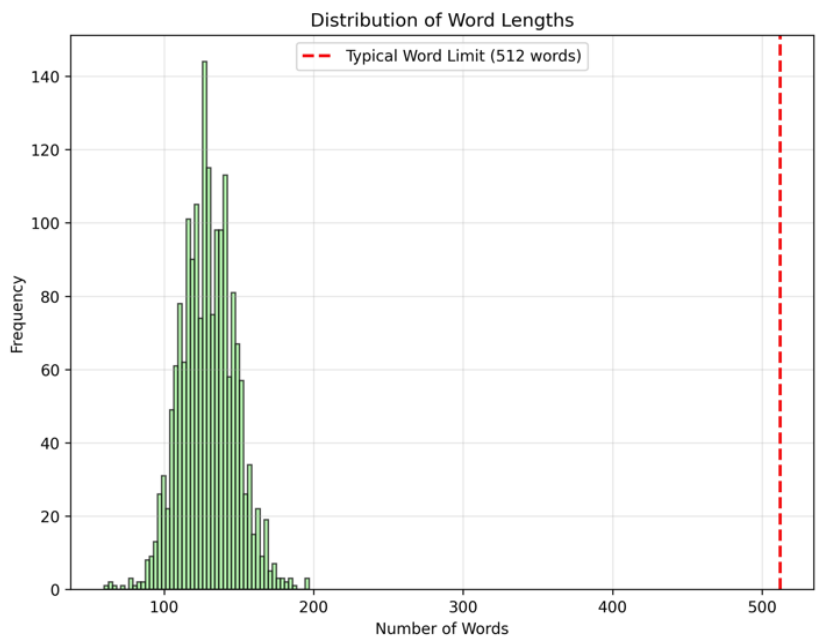

The model performance is summarized in Table 2 for essay-based scoring. When essays were used as the input data for model training, ELECTRA-large yielded the highest QWK (0.8485), deBERTa-v3-large followed while distilBERT yielded the lowest QWK (0.8111). BERT-base-uncased performed similarly to those reported in other studies (Xue et al., 2021). For scoring accuracy at each score category, only ELECTRA-large and RoBERTA-base yielded F1 score of 0.5 for score 0 while DeBARTa-base yielded F1 score of 0.3333. Other models did not assign score 0 to any essays correctly. For other score categories, ELECTRA-large yielded high F1 scores for scores of 2, 3, and 4 and moderate F1 score for score 1. The model with the highest QWK also yielded the highest Spearman rank order correlation.

Table 2. Model Performance for Essay-Based Models

| # | Model | QWK | Spearman Correlation | F1_score_0 | F1_score_1 | F1_score_2 | F1_score_3 | F1_score_4 |
|---|---|---|---|---|---|---|---|---|
| 1 | bert-base | 0.8286 | 0.8385 | 0.0000 | 0.5000 | 0.6424 | 0.7327 | 0.7486 |
| 2 | deberta-base | 0.8220 | 0.8078 | 0.3333 | 0.5634 | 0.6452 | 0.7323 | 0.6623 |
| 3 | deberta-v3-large | 0.8405 | 0.8416 | 0.0000 | 0.6216 | 0.6809 | 0.7247 | 0.7654 |
| 4 | distilbert-base | 0.8111 | 0.8060 | 0.0000 | 0.5676 | 0.6375 | 0.7455 | 0.6939 |
| 5 | electra-large | 0.8495 | 0.8485 | 0.5000 | 0.6154 | 0.7229 | 0.745 | 0.7374 |
| 6 | roberta-base | 0.8306 | 0.8207 | 0.5000 | 0.5667 | 0.7093 | 0.7492 | 0.6667 |
| 7 | roberta-large | 0.8221 | 0.8154 | 0.0000 | 0.5970 | 0.6705 | 0.7453 | 0.7248 |

Table 3 summarizes the evaluation metric for rationale-based models using GPT-4.1. DeBERTA-v3-large produced the highest QWK (0.8194), followed by deBERTa-base (0.8141) while BERT-base-uncased yielded the lowest QWK (0.7825). In general, models based on rationales generated by GPT-4.1 performed worse than models based on essays. However, a noteworthy point is that rationale-based models yielded higher F1 score for score 0, while majority of the essay-based models did not assign score 0 to any essay. F1 scores for other score categories were generally lower than those for essay-based models. BERT-base-uncased and ELECTRA-large yielded QWK below 0.8. Spearman rand-order correlation aligned with QWK.

Table 3. Model Performance for Rationale-Based Models (GPT-4.1)

| # | Model | QWK | Spearman Correlation | F1_score_0 | F1_score_1 | F1_score_2 | F1_score_3 | F1_score_4 |
|---|---|---|---|---|---|---|---|---|
| 1 | bert-base | 0.7825 | 0.7732 | 0.4000 | 0.5867 | 0.6552 | 0.7176 | 0.5000 |
| 2 | deberta-base | 0.8141 | 0.7900 | 0.5000 | 0.6400 | 0.5816 | 0.7347 | 0.6621 |
| 3 | deberta-v3-large | 0.8194 | 0.8063 | 0.5000 | 0.5152 | 0.6358 | 0.7394 | 0.6752 |
| 4 | distilbert-base | 0.8031 | 0.7788 | 0.4444 | 0.6301 | 0.6447 | 0.7434 | 0.6522 |
| 5 | electra-large | 0.7937 | 0.7761 | 0.5882 | 0.5455 | 0.6286 | 0.6688 | 0.6093 |
| 6 | roberta-base | 0.8105 | 0.7826 | 0.5882 | 0.5667 | 0.6424 | 0.7108 | 0.6301 |
| 7 | roberta-large | 0.8050 | 0.7936 | 0.3077 | 0.5753 | 0.6000 | 0.7251 | 0.6667 |

Table 4 summarizes the evaluation metric for rationale-based models using GPT-5. RoBERTa-large produced the highest QWK (0.8027), the only model with QWK above 0.8, followed by deBERTa-v3-large (0.7993) while distilBERT yielded the lowest QWK (0.7775). In general, models based on rationales generated by GPT-5 performed worse than rationales generated by GPT-4.1 and models based on essays. This might be due to the additional notes added to prompting GPT-5 in generating rationales, leading to too much succinctness. Further investigation is needed. However, it is also worth note that similar to models based on the rationales generated by GPT-4.1, GPT-5 rationale-based large models (deBERTa-v3-large, ELECTRA-large, and RoBERTA-large) all yielded higher F1 score (from 0.51 to 0.60) for score 0, which surpassed the essay-based models for scoring essays with score 0. F1 scores for other score categories were generally lower than those for rationale-based models for GPT-4.1 and essay-based models. The rank orders of Spearman correlation differed from the patterns observed above, which might be due to the fact that the top values were very close to each other.

Table 4. Model Performance for Rationale-Based Models (GPT-5)

| # | Model | QWK | Spearman Correlation | F1_score_0 | F1_score_1 | F1_score_2 | F1_score_3 | F1_score_4 |
|---|---|---|---|---|---|---|---|---|
| 1 | bert-base | 0.7909 | 0.7768 | 0.0000 | 0.6353 | 0.5965 | 0.6330 | 0.5949 |
| 2 | deberta-base | 0.7848 | 0.7759 | 0.0000 | 0.5823 | 0.6433 | 0.5818 | 0.5806 |
| 3 | deberta-v3-large | 0.7993 | 0.7612 | 0.6000 | 0.6471 | 0.6543 | 0.6790 | 0.5205 |
| 4 | distilbert-base | 0.7775 | 0.7420 | 0.4615 | 0.5902 | 0.5976 | 0.6806 | 0.5714 |
| 5 | electra-large | 0.7880 | 0.7618 | 0.5714 | 0.5667 | 0.6448 | 0.6513 | 0.5912 |
| 6 | roberta-base | 0.7870 | 0.7735 | 0.1667 | 0.6173 | 0.6303 | 0.7041 | 0.5161 |
| 7 | roberta-large | 0.8027 | 0.7762 | 0.5294 | 0.3448 | 0.6369 | 0.5414 | 0.5659 |

As essay-based models performed better overall and at high score categories with good representation while rationale-based models improved scoring accuracy for score 0 with less representation due to smaller sample sizes, ensemble models were explored. As a baseline for comparison, the ensemble models built on models trained on essays were explored first and summarized in Table 5. Stacking ensemble led to the highest QWK (0.8564), exceeding the best performing individual LM (DeBERTA-v3-large) based on essays. However, F1 score for score 0 was still below 0.5 though F1 scores for other score categories improved.

Table 5. Model Performance for Ensemble Models with Essays only

| Essay | Model | QWK | Spearman Correlation | F1_score_0 | F1_score_1 | F1_score_2 | F1_score_3 | F1_score_4 |
|---|---|---|---|---|---|---|---|---|
| 1 | Stacking Ensemble | 0.8564 | 0.8519 | 0.2000 | 0.6471 | 0.7368 | 0.7796 | 0.7468 |
| 2 | Tiered Ensemble | 0.8509 | 0.8517 | 0.3636 | 0.6154 | 0.7135 | 0.7354 | 0.7473 |
| 3 | Elite Ensemble | 0.8497 | 0.8470 | 0.3636 | 0.6154 | 0.7093 | 0.7450 | 0.7471 |
| 4 | Weighted Median | 0.8489 | 0.8455 | 0.0000 | 0.6377 | 0.7219 | 0.7653 | 0.7407 |
| 5 | QWK Optimized Ensemble | 0.8489 | 0.8455 | 0.0000 | 0.6377 | 0.7219 | 0.7653 | 0.7407 |
| 6 | Confidence Weighted | 0.8489 | 0.8455 | 0.0000 | 0.6377 | 0.7219 | 0.7653 | 0.7407 |
| 7 | Correlation Optimized | 0.8489 | 0.8455 | 0.0000 | 0.6377 | 0.7219 | 0.7653 | 0.7407 |

Table 6 summarizes the ensemble of models trained on essays and models trained on rationales from GPT-5. Tiered ensemble model led to the highest QWK (0.8583), exceeding the best performing individual LM (DeBERTA-v3-large) based on essays and the ensemble model based on essays (Stacking ensemble). Further, F1 score for score 0 is 0.5 for Tiered ensemble, improved compared with other ensemble models based on essays only. F1 scores for other score categories were about the same as those for Stacking ensemble based on essays only.

Table 6. Model Performance for Ensemble Models with Essays and Rationales from GPT-5

| Essay | Model | QWK | Spearman Correlation | F1_score_0 | F1_score_1 | F1_score_2 | F1_score_3 | F1_score_4 |
|---|---|---|---|---|---|---|---|---|
| 1 | Tiered Ensemble | 0.8583 | 0.8564 | 0.5000 | 0.6462 | 0.7326 | 0.7405 | 0.7473 |
| 2 | Stacking Ensemble | 0.8569 | 0.8503 | 0.3636 | 0.6567 | 0.7586 | 0.7756 | 0.7179 |
| 3 | Elite Ensemble | 0.8537 | 0.8495 | 0.3636 | 0.6364 | 0.7326 | 0.7533 | 0.7368 |
| 4 | Weighted Median | 0.8461 | 0.8381 | 0.2000 | 0.6571 | 0.7368 | 0.7508 | 0.7000 |
| 5 | QWK Optimized Ensemble | 0.8412 | 0.8289 | 0.2000 | 0.6761 | 0.7262 | 0.7412 | 0.6835 |
| 6 | Correlation Optimized | 0.8412 | 0.8289 | 0.2000 | 0.6761 | 0.7262 | 0.7412 | 0.6835 |
| 7 | Confidence Weighted | 0.8409 | 0.8299 | 0.2000 | 0.6761 | 0.7283 | 0.7320 | 0.6875 |

Table 7 summarizes the ensemble of models trained on essays and models trained on rationales from GPT-4.1. Stacking ensemble model led to the highest QWK (0.8659), exceeding the best performing individual LM (DeBERTA-v3-large) based on essays and the ensemble model based on essays and rationales from GPT-5 (Tiered ensemble). Further, F1 score for score 0 was 0.5 for Tiered ensemble and 0.3636 for other ensemble models. F1 scores for other score categories improved over other ensemble models.

Table 7. Model Performance for Ensemble Models with Essays and Rationales from GPT-4.1

| Essay | Model | QWK | Spearman Correlation | F1_score_0 | F1_score_1 | F1_score_2 | F1_score_3 | F1_score_4 |
|---|---|---|---|---|---|---|---|---|
| 1 | Stacking Ensemble | 0.8659 | 0.8598 | 0.3636 | 0.6765 | 0.7674 | 0.7937 | 0.7403 |
| 2 | Tiered Ensemble | 0.8616 | 0.8589 | 0.5000 | 0.6667 | 0.7412 | 0.7448 | 0.7473 |
| 3 | Elite Ensemble | 0.8578 | 0.8535 | 0.3636 | 0.6567 | 0.7412 | 0.7559 | 0.7399 |
| 4 | Correlation Optimized | 0.8555 | 0.8504 | 0.3636 | 0.6286 | 0.7273 | 0.7938 | 0.7248 |
| 5 | QWK Optimized Ensemble | 0.8555 | 0.8504 | 0.3636 | 0.6286 | 0.7273 | 0.7938 | 0.7248 |
| 6 | Weighted Median | 0.8543 | 0.8488 | 0.3636 | 0.6286 | 0.7273 | 0.7901 | 0.7200 |
| 7 | Confidence Weighted | 0.8491 | 0.8449 | 0.3333 | 0.5882 | 0.7262 | 0.7760 | 0.7097 |

Table 8. Model Performance for Ensemble Models with Essays and both Rationales

| Essay | Model | QWK | Spearman Correlation | F1_score_0 | F1_score_1 | F1_score_2 | F1_score_3 | F1_score_4 |
|---|---|---|---|---|---|---|---|---|
| 1 | Stacking Ensemble | 0.8703 | 0.8599 | 0.6154 | 0.6970 | 0.7701 | 0.7937 | 0.7368 |
| 2 | Elite Ensemble | 0.8644 | 0.8616 | 0.3636 | 0.6567 | 0.7657 | 0.7718 | 0.7456 |
| 3 | Tiered Ensemble | 0.8615 | 0.8590 | 0.3636 | 0.6567 | 0.7657 | 0.7577 | 0.7356 |
| 4 | Weighted Median | 0.8437 | 0.8314 | 0.3636 | 0.6667 | 0.7412 | 0.7492 | 0.6710 |
| 5 | QWK Optimized Ensemble | 0.8408 | 0.8271 | 0.3636 | 0.6667 | 0.7381 | 0.7500 | 0.6579 |
| 6 | Confidence Weighted | 0.8374 | 0.8229 | 0.3636 | 0.6667 | 0.7337 | 0.7438 | 0.6490 |
| 7 | Correlation Optimized | 0.8369 | 0.8222 | 0.3636 | 0.6667 | 0.7381 | 0.7453 | 0.6400 |

Table 8 summarizes the ensemble models based on all models (21 models). Stacking ensemble model led to the highest QWK (0.8703), the best among all trained models in this study. Further, F1 score for score 0 increased to 0.6154, the best among all models. F1 scores for other score categories improved slightly or remained the same as other best performing models. For all the above ensemble models, Spearman correlations aligned with the ranking of QWK for each model. For the current ensemble model over all input data, Elite ensemble yielded slightly higher Spearman correlation compared with Stacking ensemble with the highest QWK.

For scoring with GPT, GPT-4.1 yielded a QWK of 0.6067 while GPT-5 yielded a QWK of 0.7283. No studies have explored GPT-4.1 and GPT-5 for essay scoring. These values are comparable with those reported in other studies (see Appendix A) for GPT-4 (Xiao, et al, 2024; 0.7284) and ChatGPT (Jang et al., 2025; 0.606).

## Summary and Discussion

This study evaluates the utility of scoring rationales generated by GPT in enhancing scoring accuracy. The fine-tuned encoder LMs with rationales did not perform better than encoder-only LMs based on essays. However, LMs fine-tuned on rationales seemed to provide auxiliary information about scoring accuracy at individual score level. Our best performing ensemble model yielded the highest QWK of 0.87 for Prompt 6 in the ASAP essays. To our knowledge, this might be the highest QWK reported in public domain for ASAP Prompt 6. Ensemble modeling shows that these models contain complementary information that can be optimally combined to improve scoring accuracy at individual score level and ultimately the overall scoring accuracy. The findings of our study provided empirical evidence related to augmenting data for automated essay scoring with rationales generated by LLMs.

We also trained supervised machine learning models (X-gradient boosting and light gradient boosting) on embeddings from the best performing LMs (ELECTRA and DeBERTa) based on essays and rationales respectively. QWKs for models trained on embeddings, either from essays or rationales, or combined, yielded QWK around 0.72 to 0.78, which did not surpass those for LMs. As ELECTRA and DeBERTa are not optimized for semantic similarity, they work for classification after fine-tuning but could be mediocre for embeddings. Thus, high-quality sentence embeddings from Sentence-Transformers families such as all-MiniLM and E5 can be explored in future studies.

As clearly documented in score distribution, class imbalance is present in this dataset. Score 0 and 1 are less represented in the dataset. This current study did not apply strategies to rebalance score representation via data augmentation methods. Future exploration can explicitly implement data augmentation methods to improve scoring accuracy with LMs based on essays and rationales respectively. It is expected that data augmentation will improve essay scoring accuracy (e.g., Jiao, Lnu, & Zhai, 2024).

We experimented with different prompting strategies with GPT-4.1 and found that our new prompting increased QWK from 0.6067 to 0.6489. We will incorporate different prompting strategies in future exploration. Further, our current prompting to GPT-5 for rationale generation might be too strict. We will experiment with other options that GPT-5 rationale can be more elaborated.

This study demonstrated the utilities of rationales generated by GPT-4.1 and GPT-5 in enhancing automated scoring accuracy based on essays only. As data augmentation, rationales generated by LLMs may provide information complementary to that embedded in essays. One of the most important findings worthy of highlights is that LLM rationales helped to increase the scoring accuracy for scores with less representation. This echoes what is observed in item response theory modeling related to the utility of items with low discrimination parameters, when such items may provide more item information for certain ability range compared with items with high discrimination parameters. Rationale-based scoring models in general may not lead to scoring accuracy comparable to essay-based scoring models. However, rationale-based scoring may complement essay-based scoring to enhance the overall scoring accuracy and the scoring accuracy at specific scores.

# Appendix A

Table A.1. Example Studies Using the ASAP Prompt 6 for Automated Essay Scoring

| Authors | Models | Features | QWK |
|---|---|---|---|
| Shemis (2014) | Human-Human | Human intelligence | 0.776 |
| Mahana et al. (2012) | linear regression with a polynomial basis function | Bag of words, numerical features, parts of speech count, orthography, structure and organization | 0.700 |
| Taghipour and Ng (2016) | LSTM | Length, parts-of-speech, word overlap with the prompt, Bag of n-grams | 0.813 |
| Liang et al. (2017) | Siamese Bidirectional LSTM | GloVe 50-dimension word embedding | 0.820 |
| Dong et al. (2017) | Attention-based with LSTM +CNN | Word embedding, content-based features | 0.811 |
| Zhao et al. (2017) | MN: memory-augmented neural model | Word embedding, position encoding | 0.830 |
| Wang et al. (2018) | reinforcement learning | word embedding via word2vec | 0.798 |
| Rodriguez et al. (2019) | Ease | Traditional BOW | 0.771 |
| Rodriguez et al. (2019) | BERT (base) | | 0.805 |
| Rodriguez et al. (2019) | LSTM | BOW | 0.813 |
| Rodriguez et al. (2019) | BERT Ensemble | | 0.815 |
| Rodriguez et al. (2019) | XLNet Ensemble | BOW | 0.805 |
| Rodriguez et al. (2019) | BERT+ XLNet Ensemble | | 0.815 |
| Rodriguez et al. (2019) | BERT+CNN Ensemble | | 0.819 |
| Yang et al. (2020) | $R^2$ BERT | | 0.847 |
| Muangkammuen and Fukumoto (2020) | multi-task with word and sentence sentiment | | 0.816 |
| Mathias et al (2020) | | | 0.821 |
| Uto and Uchida (2020) | BERT + Features | | 0.817 |
| Uto, Xie, and Ueno (2020) | BERT+essay-level features | Essay-level features | 0.817 |
| Jeon and Strube (2021) | | | 0.820 |
| Xue et al. (2021) | BERT finetune | | 0.826 |
| Xue et al. (2021) | hierarchical-BERT | | 0.832 |
| Ormerod et al. (2021) | BERT base | | 0.805 |
| Ormerod et al. (2021) | Albert base | | 0.816 |
| Ormerod et al. (2021) | Albert large | | 0.807 |
| Ormerod et al. (2021) | Electra small | | 0.787 |
| Ormerod et al. (2021) | Mobile BERT | | 0.808 |
| Ormerod et al. (2021) | Reformer | | 0.762 |
| Ormerod et al. (2021) | Electra+Mobile BERT | | 0.802 |
| Ormerod et al. (2021) | Efficient Ensemble | | 0.822 |
| Ormerod (2022) | DeBERTa Large | | 0.834 |

| Yao and Jiao (2023) | Gradient Boosting | features with readability measures | 0.724 |
| Cho et al. (2023) | BERT with data augmentation | | 0.839 |
| Xiao et al. (2024) | GPT-4 few-shot, with rubric | | 0.7284 |
| Xiao et al. (2024) | Fine-tuned GPT-3.5 | | <span style="color:red">0.848</span> |
| Xiao et al. (2024) | Fine-tuned Llama3 | | 0.7489 |
| Xiao et al. (2024) | GPT with their fast module | | 0.7753 |
| Shermis (2024) | GPT-4o | | 0.780 |
| Lee et al (2024) | ChatGPT | | 0.668 |
| Lee et al (2024) | Llama2-7B | | 0.507 |
| Lee et al (2024) | Llama2-13B | | 0.590 |
| Lee et al (2024) | Mistral-7B-instruct | | 0.657 |
| Jang et al (2025) | ChatGPT | | 0.606 |
| Jang et al (2025) | Claude | | 0.630 |
| Chu et al (2025) | T5 | | 0.776 |
| Chu et al (2025) | Flan-T5 with GPT rationale | | 0.762 |
| Chu et al (2025) | BART with Llama rationale | | 0.748 |
| Chu et al (2025) | Pegasus with Llama rationale | | 0.714 |
| Chu et al (2025) | LED with Llama rationale | | 0.745 |
| Shibata & Miyamura (2025) | GPT-4o-mini-Pairwise comparison | | 0.783 |
| Shibata & Miyamura (2025) | Mistral-7B-Pairwise comparison | | 0.792 |
| Shibata & Miyamura (2025) | Llama 3.2-3B-Pairwise comparison | | 0.756 |
| Shibata & Miyamura (2025) | Llama 3.1-8B-Pairwise comparison | | 0.707 |
| Shibata & Miyamura (2025) | GPT-4o-Pairwise comparison | | 0.766 |

Appendix B

**LLM Prompting**

```
TASK
You are an experienced essay grader. Score the following essay
holistically using the provided rubric.

READING PASSAGE:
{passage}

ESSAY PROMPT:
{prompt}

STUDENT ESSAY:
{essay}

SCORING RUBRIC (Holistic - Single Score from 0 to 4):
{rubric_text}

INSTRUCTIONS:
1. Read the passage, prompt, and student essay carefully
2. Evaluate the essay holistically against the rubric
3. Assign ONE score from 0 to 4
4. Provide a detailed rationale explaining why this score was assigned
5. Reference specific elements from the essay in your rationale but do
not repeat the rubric

Additional instruction (for GPT-5 only)
6. Keep the rationale focused and avoid unnecessary verbosity
7. Use direct, clear language without excessive elaboration
8. Focus on the key strengths and weaknesses that determined the score
9. Each rationale should Not be more than 512 tokens

Scoring Notes (for GPT-4.1 only)
The obstacles to dirigible docking include:
1. Building a mast on top of the building
2. Meeting with engineers and dirigible engineers
3. Transmitting the stress of the dirigible all the way down the
building; the frame had to be shored up to the tune of $60,000
4. Housing the winches and other docking equipment
5. Dealing with flammable gases
6. Handling the violent air currents at the top of the building
7. Confronting laws banning airships from the area
8. Getting close enough to the building without puncturing

Other explanations will be accepted if supported by relevant evidence
from the text.
```

Please respond in the following format:
SCORE: [0-4]
RATIONALE: [Detailed explanation of why this score was assigned, with specific references to the essay content and how it aligns with the rubric criteria]

RUBRIC
4: The response is a clear, complete, and accurate description of the obstacles the builders of the Empire State Building faced in attempting to allow dirigibles to dock there. The response includes relevant and specific information from the excerpt.
 3: The response is a mostly clear, complete, and accurate description of the obstacles the builders of the Empire State Building faced in attempting to allow dirigibles to dock there. The response includes relevant but often general information from the excerpt.
 2: The response is a partial description of the obstacles the builders of the Empire State Building faced in attempting to allow dirigibles to dock there. The response includes limited information from the excerpt and may include misinterpretations.
1: The response is a minimal description of the obstacles the builders of the Empire State Building faced in attempting to allow dirigibles to dock there. The response includes little or no information from the excerpt and may include misinterpretations. OR The response relates minimally to the task.
0: The response is totally incorrect or irrelevant or contains insufficient evidence to demonstrate comprehension.